\documentclass{article}

\usepackage[nonatbib,final]{nips_2016}

\usepackage[utf8]{inputenc} 
\usepackage[T1]{fontenc}    
\usepackage{hyperref}       
\usepackage{url}            
\usepackage{booktabs}       
\usepackage{amsfonts}       
\usepackage{nicefrac}       
\usepackage{microtype}      

\usepackage{multirow}
\usepackage{epsfig}
\usepackage[flushleft]{threeparttable}
\usepackage{algorithmic}
\usepackage{algorithm}
\usepackage{bm}
\usepackage{amsmath}
\usepackage[tight,footnotesize]{subfigure}

\def\bA{{\mathbf A}}
\def\bB{{\mathbf B}}

\def\bI{{\mathbf I}}
\def\bO{{\mathbf O}}
\def\bP{{\mathbf P}}
\def\bR{{\mathbf R}}
\def\bS{{\mathbf S}}
\def\bT{{\mathbf T}}
\def\bU{{\mathbf U}}
\def\bV{{\mathbf V}}

\def\cD{{\mathcal D}}

\title{Partially Observable Markov Decision Process for Recommender Systems}

\author{
  Zhongqi Lu, Qiang Yang \\
  Hong Kong University of Science and Technology, Hong Kong \\
  \texttt{\{zluab,qyang\}@cse.ust.hk} \\
}

\begin{document}

\maketitle

\begin{abstract}
We report the `Recurrent Deterioration' (RD) phenomenon observed in online recommender systems. The RD phenomenon is reflected by the trend of performance degradation when the recommendation model is always trained based on users' feedbacks of the previous recommendations. There are several reasons for the recommender systems to encounter the RD phenomenon, including the lack of negative training data and the evolution of users' interests, etc. Motivated to tackle the problems causing the RD phenomenon, we propose the POMDP-Rec framework, which is a neural-optimized Partially Observable Markov Decision Process algorithm for recommender systems.
We show that the POMDP-Rec framework effectively uses the accumulated historical data from real-world recommender systems and automatically achieves comparable results with those models fine-tuned exhaustively by domain exports on public datasets.
\end{abstract}



\section{Introduction}\label{section:intro}
Recommender systems are widely applicable in real-world scenarios such as e-learning, online shopping, mobile app stores, etc. In recommender systems, the training data usually come from users' {\em positive} feedbacks (such as ``clicks'') towards its previous recommendations. This process goes on recurrently in the recommender system. Under this recurrent ``train with feedbacks, make recommendations, collect feedbacks, re-train, \ldots'' mechanism, the outputs of the previous recommendation models would affect the input of the future training and the pool of recommendation candidates would become smaller when the recurrences go on. Therefore, this specific characteristic of online recommender systems could lead to a significant performance degradation as the system runs. This phenomenon is called `Recurrent Deterioration' (RD).

In our analysis, the recommender systems encounters the RD phenomenon mainly because of two reasons: (a) potential system biases caused by a lack of negative training samples (For example, the advertising recommender systems only see the click as a positive feedback), and the biases accumulate as the recurrent recommendation mechanism runs, and (b) frequent changes of data distributions caused by the changes of users' interests and the changes of recommendation context. Some tweaks, such as sampling for negative samples~\cite{marlin2009collaborative,hernandez2014probabilistic} and/or neighborhood regularization~\cite{melville2002content}, could make the RD phenomenon less severe. But the two problems are not solved naturally. Besides, these solutions need lots of human efforts.

In the previous researches, modeling the sequential recommendation process based on the Markov Decision Process (MDP) was proposed~\cite{shani2005mdp}. However, because the optimization of the MDP assumes full observations of its states and thus requires dense training data, the MDP approach is not very practical for current online recommender systems, in which an user usually views less than $1\%$ items.
The sparse nature of the training data in recommender systems indicates the observations of users' behaviours shall be ``partial''. Therefore, we propose to use a Partially Observable Markov Decision Process (POMDP) to model the sequential recommendations and tackle the RD phenomenon.
We call our POMDP based framework for recommender systems ``POMDP-Rec''.
Intuitively, we consider a recommender system as an agent exploring an environment. The environment could be observed through the users' positive feedbacks. Based on limited observations, the recommender system estimates the interests of its users. In other words, the recommender system (as an agent) has some hidden belief about what state it is in, and what would be the next state to be expected. Whenever taking some actions (i.e. recommendations), the recommender system gets its rewards based on short-term user feedbacks (such as clicks) and/or long-term outcomes (such as users' payments). The objective is to maximize the cumulative rewards.


Our proposed POMDP-Rec framework is capable to tackle the two problems causing RD phenomenon, i.e., a lack of negative samples in recurrently training a recommender system, and frequent change of the data distribution. Different from classical supervised learning methods for recommender systems, this training process of the POMDP-Rec avoids the needs of negative sample. Indeed, the classical supervised learning methods try to learn a ranking of all the candidates before making the recommendations, and thus need to compare with positive and negative samples. But for this POMDP-Rec framework, the recommender system will only need to train on recommendations with positive feedbacks. Another obvious benefit from the POMDP-Rec modeling is the ability to capture the shifts of data distributions caused by users' changing interests and the changing recommendation context, because it is the transitions between states that are considered in the POMDP-Rec modeling.

Other than solving the two problems causing RD phenomenon, the POMDP-Rec framework is suitable for modeling the recommender systems because it can use the historical training data in the recommender systems efficiently. Due to the Markovian property, we could train the POMDP-Rec without the need to maintain the whole sequential order of user's behaviours. Besides, because our POMDP-Rec model does not have any assumption on user's future interests and the training could be done off-line, we could use as many historical samples as possible without worrying about the changing of users' interests during the testing. So far as we are concerned, this is the first literature to introduce the POMDP-based modeling for recommender systems.


\section{Preliminaries}
\subsection{The POMDP}
The recommendation problem considered in this paper could be formulated by a POMDP~\cite{smallwood1973optimal}. A POMDP models a decision process in which it is assumed that the system dynamics are determined by an MDP, but the underlying state cannot be directly observed. It maintains a probability distribution over the set of possible states, based on a set of observations and observation probabilities, as well as the underlying MDP.

The POMDP can be a proper modeling of the recommender systems because of two reasons. On one hand, the Markovian assumption has been proven to be valid for the sequential recommendations~\cite{shani2005mdp}. On the other hand, we shall not expect to fully observe an user's actions on all items in a recommender system. Therefore, we shall avoid to define the users' actions as states. Instead, we argue that the observed user's actions would depend on some hidden states with a probability. Under such a modeling, the POMDP would be suitable for the sequential recommendations.


\subsection{Neural fitted Q-learning}
Reinforcement learning methods, such as Q-learning, are often adopted to optimize the MDP-based models. In classical Q-learning~\cite{sutton1998reinforcement}, the goal is to learn a Q-value function, which is a mapping from the state and action space to an evaluation space. This Q-value function follows an important identity known as {\em Bellman equation}. This is based on an intuition: if the optimal $Q^\star (s^\prime, a^\prime)$ of the sequence $s^\prime$ at the next time-step was known for all possible actions $a^\prime$, then the optimal strategy is to select the action $a^\prime$ maximizing the expected value of $r(s,a) + \gamma \max_{a^\prime} Q^\star (s^\prime, a^\prime)$. The update rule is given by
\begin{eqnarray}
Q^\star (s, a) = r(s,a) + \gamma \max_{a^\prime} Q^\star (s^\prime, a^\prime),
\end{eqnarray}
where $s \in \bS$ denotes the state where the transition starts, $a \in \bA$ is the current action, and $s^\prime \in \bS$ is the resulting state. $r(s,a)$ is an immediate reward function $r:\bS \times \bA \rightarrow \bR$. $\gamma$ is a discounting factor.

Recent researches have demonstrated the success of using neural fitted function to approximate the Q-values in real-world applications, such as playing Atari~\cite{mnih2013playing}. As mentioned in~\cite{riedmiller2005neural}, the above process of generating Q-values can be parameterized via an instance of the Fitted Q Iteration family of algorithms~\cite{ernst2005tree}. For instance, a neural network function approximator with parameters $\theta$ could be used to estimate the Q-value. To learn $\theta$, the common way is to minimize a sequence of loss functions $L_i (\theta_i)$ iteratively,
\begin{eqnarray} \label{eq:lossfunc}
L_i (\theta_i) = ( y_i - Q(s,a;\theta_i) )^2
\end{eqnarray}
where $y_i = r(s,a) + \gamma \max_{a^\prime} Q(s^\prime, a^\prime; \theta_{i-1})$ is the target for iteration $i$. We would like to fix the parameters $\theta_{i-1}$ when optimizing the loss function $L_i$.

In this work, we would like to adopt the idea of Fitted Q Iteration procedures to estimate the Q-values and come up with a neural fitted Q-learning solution for recommender systems.

\section{POMDP-Rec for Recommender Systems}
\subsection{The POMDP-Rec framework}


Given the Markovian properties of recommender systems~\cite{shani2005mdp}, we will find the POMDP modeling to be suitable to explain the nature of recommender systems. The recommender system would not be able to fully observe the users' actions towards the entire item set, because the users usually provide limited feedbacks~\cite{adomavicius2005toward}.

Formally, the POMDP-Rec framework in this paper is represented by an $8$-tuple $(\bS, \bA, \bT, \bR, \Omega, \bO, \bP, \gamma)$, where
\begin{itemize}
\item $\bS$ is a set of hidden states.
\item $\bA$ is a set of actions. For example, the recommendations.
\item $\bT(s^\prime \mid s, a)$ is a set of conditional transition probabilities when taking action $a \in \bA$, which causes the environment transit from state $s$ to $s^\prime$.
\item $\bR : \bS \times \bA \longrightarrow \bR$ is the reward function.
\item $\bO$ is a set of observations of users' behaviours.
\item $\Omega(o \mid s, a)$ is a set of conditional observation probabilities of receiving an observation $o \in \bO$ after the agent takes action $a$ at hidden state $s$.
\item $\bP$ is a probability distribution, from which the underlying state $s \in \bS$ is generated.
\item $\gamma \in [0, 1]$ is the discount factor.
\end{itemize}

Under the framework of POMDP-Rec, at each time point, the recommender system (agent) interacts with the environment and gets an observation $o \in \bO$ from users' (positive) feedbacks. Based on the observation $o$, the underlying state $s \in \bS$ is then generated from the probability distribution $\bP$. At state $s$, an action $a \in \bA$ is chosen to determine both the reward $r = \bR(s, a)$ and the next state $s^\prime \sim \bT(s,a)$.

\subsection{Belief Approximation of POMDP-Rec}
Although the POMDP approach provides a natural way to model the recommender systems, solving it is non-trivial~\cite{hauskrecht2000value}. To solve for the proposed POMDP-Rec framework under the recommender system settings, we developed an approximation by introducing the beliefs of the hidden states in POMDP, known as belief states. To estimate the belief states, we propose to refer to the low-dimensional factor model~\cite{mnih2007probabilistic}. The approach could handle the sparseness of the observations, because the belief states are learned in a collaborative manner.

In the low-dimensional factor model, a user's actions towards an item is modeled by linearly combining the estimations of items' latent features and the estimations of user's latent interests.
Following~\cite{mnih2007probabilistic}, the estimations of items' latent features $\bV$ and the estimations of user's latent interests $\bU$ are assumed to be from some particular normal distributions with zero-mean and corresponding variances. With a partially observed user-item matrix $\bO$, $\bV$ and $\bU$ are inferred from:
\begin{eqnarray}
p(\bO \mid \bU, \bV, \sigma^2) = \prod_{i=1} \prod_{j=1} \left(\mathcal{N}(\bO_{ij} \mid {\bU_i}^\top \bV_j, \sigma^2)\right)^{\phi_{ij}}
\end{eqnarray}
where $\mathcal{N}(x \mid \mu, \sigma^2)$ is the probability density function of the Gaussian distribution with mean $\mu$ and variance $\sigma^2$, and $\phi_{ij} \in \{0, 1\}$ is a indicator of observing $(i ,j)$, and
\begin{eqnarray} \label{eq:belief}
p\left(\bU \mid \sigma_u^2\right) = \prod_{i=1} \mathcal{N}\left(\bU_i \mid 0, \sigma_u^2 \bI_k \right), \qquad p\left(\bV \mid \sigma_v^2\right) = \prod_{j=1} \mathcal{N}\left(\bV_j \mid 0, \sigma_v^2 \bI_k \right),
\end{eqnarray}
where $\bU_i$ is the estimation of user $i$'s latent interests, and $\bV_j$ is the estimation of item $j$'s latent features.
The belief state $b_{ij}$ (corresponding to the interaction between user $i$ and item $j$) is defined as the concatenation of the estimation of user's latent interests and the estimation of item's latent features, and is denoted as $b_{ij} = \langle \bU_i, \bV_j \rangle$, where $\bU_i$ and $\bV_j$ are as in Equation~\ref{eq:belief}.

To infer user $i$'s interaction on item $j$, we split $b_{ij}$ half by half so that the first half is for user's latent interests $\bU_i$ and the rest is for item's latent features $\bV_j$. The prediction is give by $h(b_{ij}) = \bU_i^T \bV_j$.

With the definition of belief states, solving the original POMDP-Rec is equivalent to solving an MDP, which is defined as a tuple $(\bB, \bA, \tau, r, \gamma)$, where
\begin{itemize}
\item $\bB$ is the set of belief states. $\bB = \{ \langle \bU_i, \bV_j \rangle \}_{i=1,2,\ldots,\mid \bU \mid ; j = 1,2,\ldots, \mid \bV \mid}$, where $\bU$ and $\bV$ come from Equation~(\ref{eq:belief}),
\item $\bA$ is the same set of actions as for the original POMDP,
\item $\tau$ is the belief state transition function,
\item $\bR : \bB \times \bA \longrightarrow \bR$ is the reward function on belief states,
\item $\gamma$ is the discount factor, same to the one in the original POMDP.
\end{itemize}

The above transformation introduces a Markovian belief state that allows the POMDP-Rec to be re-formulated as a solvable MDP where every belief is defined as a state. An important consequence is that the belief states form a fully observable MDP, and thus it is a {\em sufficient statistic} for choosing optimal actions~\cite{Bertsekas:1987:DPD:26970}.

Q-learning has been adopted in solving the POMDP problems in some recent works~\cite{hausknecht2015deep,zhao2014exploration}. Because the future recommendations depend on how the user's behaviours change, we would like to use the transitions of the belief states as the inputs of the Q function. Details are given in the next section.

\subsection{Solving POMDP-Rec for Recommendations}
We would like to solve the POMDP-Rec by Q-learning~\cite{watkins1992q}. As the input of the Q function, we define a transition of belief states $\langle b,b^{\prime} \rangle$, where $b^\prime$ is the consecutive belief state after $b$. The Q-value function is the solution to the Bellman optimality equation:
\begin{eqnarray}\label{eq:bellman}
Q(\langle b,b^{\prime} \rangle) = r + \gamma \sum_{b^{\prime\prime} \in \bB} \tau(b^{\prime\prime} \mid b^\prime) Q(\langle b^\prime,b^{\prime\prime} \rangle)
\end{eqnarray}

To handle the complex real-world scenarios, we introduce a neural network to approximate the value function $Q$. Our approach is an instance of the Fitted Q Iteration family~\cite{ernst2005tree}, which has successful applications in playing Atari games~\cite{mnih2013playing}.

The POMDP-Rec framework consists of two major steps: $(1)$ Generating a training pattern set $PS$, and $(2)$ training these patterns within a neural network. Besides, we found in experiments that randomly shuffling the training pattern set $PS$ would lead to better results. The algorithm is shown in Algorithm~\ref{algorithm:dqn}. We provide the key implementation details in Section~\ref{section:imp_details}.

The learned Q function is used to evaluate future recommendations. In order to make future recommendations, we generate the candidates by varying the parameters of a matrix factorization model~\cite{zhuang2013fast}, and the candidate with the highest Q value is adopted for the future recommendation.

\begin{algorithm}[tb]
\caption{\footnotesize POMDP solver for Recommender Systems}
\begin{algorithmic}

\STATE {\bf{Input}:} set of transitive observations $\cD = \{ (o_{ij}; a; o^\prime_{ij}), i\in \{1,\ldots, \mid \bU \mid \}, j \in \{1,\ldots, \mid \bV \mid \} \}$; number of iterations $N$;
\STATE {\bf{Output}:} a neural network to estimate the Q-value function $Q_{N}$;

\STATE
\STATE {$iter = 0$; \%Init iteration counter}
\STATE {Init\_MLP()$\rightarrow Q_0$; \%Rand init the neural network}
\STATE {Estimate a transition function $\tau(.)$ by sampling from $\cD$ and calculating belief states;}
\WHILE {$iter < N$}
    \STATE Generate pattern set $PS = \{ (input^{ij}, target^{ij}) \}$ by iterating over the transitive observations  $\{ (o_{ij}; a; o^\prime_{ij}) \}$, where the sub-steps are:
    \STATE\hspace{\algorithmicindent}\hspace{\algorithmicindent} Generate $b_{ij} = \langle \bU_i, \bV_j \rangle$ from Eq.~\ref{eq:belief} with $\bO$;
    \STATE\hspace{\algorithmicindent}\hspace{\algorithmicindent} Generate $b^\prime_{ij} = \langle \bU^\prime_i, \bV^\prime_j \rangle$ from Eq.~\ref{eq:belief} with $\bO^\prime$;
    \STATE\hspace{\algorithmicindent}\hspace{\algorithmicindent} Calculate reward $r(b^\prime_{ij}, a)$ by Eq.~\ref{eq:reward};
    \STATE\hspace{\algorithmicindent}\hspace{\algorithmicindent} $input^{ij} = \langle b_{ij}, b^\prime_{ij} \rangle$;
    \STATE\hspace{\algorithmicindent}\hspace{\algorithmicindent} $target^{ij} = r(b^\prime_{ij}, a) + \gamma \sum_{b^{\prime\prime} \in \bB} \tau(b^{\prime\prime} \mid b_{ij}^\prime) Q(\langle b_{ij}^\prime,b^{\prime\prime} \rangle )$;  
    \STATE Randomly shuffle pattern set $PS$;
    \STATE Train\_NN($PS$) $\rightarrow Q_{iter+1}$ by optimizing Eq.~\ref{eq:lossfunc} using SGD;
    \STATE $iter := iter + 1$;
\ENDWHILE

\STATE

\end{algorithmic}
\label{algorithm:dqn}
\end{algorithm}

\subsection{Properties of POMDP-Rec}
The POMDP-Rec framework is naturally fitted for the training data from recommender system. The training data collected in recommender systems have three main characteristics.

First, the users' feedbacks are usually unevenly distributed. For instance, an online advertising recommender system only see positive feedbacks. But no feedback may indicate either the negative feedback or the users miss the recommendations~\cite{pan2008one}. The lack of negative training samples may cause a big problem for the supervised learning methods, whose learning objective is essentially classifying the positive and negative feedbacks. But it would not affect our proposed POMDP-Rec, because the objective of the POMDP-Rec is to make the recommendations that maximize the prediction accuracy for positive feedbacks.

Secondly, training data in recommender systems could have been accumulated for a long time span. The popular Collaborative Filtering (CF) approaches for recommender systems aim to capture the users' up-to-date interests, thus the historical data are usually used with a time decay factor~\cite{koren2009collaborative} and lots of data could be ``expired''. For our POMDP-Rec framework, because we are focusing on the transitions of belief states, instead of the users' interests, thus all historical feedbacks could be used, which leads to better data efficiency.

Thirdly, the data are sparse at each time interval. Handling the sparseness in data from recommender systems has been a popular research direction. Our partially observable assumption comes right from this sparseness nature of data in recommender systems. And we handle the sparse observation by collaboratively learning the belief states.

\subsection{Implementation details} \label{section:imp_details}

\noindent {\bf Preprocess data for training}
The raw data accumulated in a recommender system can be collected in triples of the form $(o, a , o^\prime)$, where $o$ is the original observation, $a$ is the recommendation to be made, and $o^\prime$ is the next observation. The consecutive observations $o$ and $o^\prime$ shall be related to the same (user, item) pair. The set of these triples is denoted as the sample set $\cD$.

\noindent {\bf Definition of rewards}
The definition of rewards is important to the Q function. By inheriting the notations in the previous sections, we introduce a definition of the reward:
\begin{eqnarray}\label{eq:reward}
r(b^\prime,a) = \frac{1}{1+\exp(C * ( \sqrt{ \sum_{i=1}^m \sum_{j=1}^n I_{ij} (a(i,j) - h(b^\prime_{ij}) )^2 / \mid I \mid} ) )}
\end{eqnarray}
where $h(b^\prime_{ij})$ is the future interaction between user $i$ and item $j$ inferred from the belief $b^\prime_{ij}$ and it serves as the ground truth, $I_{ij}$ is the $0/1$ indicator function for the availability of $a(i,j)$, and $C$ is a scaling constant to be set according to the scales of the input data, e.g., $C=0.1$ for a $0-100$ rating prediction problem.

There could be various choices of the reward functions, depending on the applications. For example, if we would like to optimize the Click Through Rate (CTR) for the recommender system, the reward could be defined as the real CTR. Directly optimizing the future reward is another advantage of our POMDP-Rec framework.

\noindent {\bf Transition function}
The transition function $\tau(.)$ is essential to estimate the expected value function as in Equation~(\ref{eq:bellman}). In our experiments, we learn a logistic regression function $y=f(b, b^\prime)$, where $y=1$ if there exists a transition from belief state $b$ to $b^\prime$. After learning the regression function $f(\cdot,\cdot)$, we use its output value to estimate the transition probability.

\noindent {\bf Implementation of NN} The approximation algorithm for the value function is not the main concern of this work. Thus, we use the regular implementation of a neural network to estimate the value function. The neural network is composed of an input layer, one hidden layer, and an output layer.

\section{Experiments}

\subsection{Dataset and setup}
We would like to test our proposed POMDP-Rec framework on the public datasets that are retrieved from real world recommender systems. From real world recommender systems, the data are likely to (a) be sparse, (b) contain uneven distributions of numerical ratings, and (c) demonstrate users' changing interests if the system runs for a relatively long time. We found the MovieLens dataset and the Yahoo Music dataset to be ideal as an off-line dataset to test the POMDP-Rec framework and the state-of-the-art methods, although the anonymilization of these public datasets prevent us from further analyzing the belief states, and Q function etc in the POMDP-Rec model.

The $1$M MovieLens dataset~\cite{harper2015movielens} contains about 1 million ratings of 3,952 movies by 6,040 users with Unix timestamps. Each user has at least $20$ ratings, and the average sparsity of the whole dataset is about $4.2$\%. The MovieLens dataset was collected from the MovieLens web site,\footnote{\url{http://movielens.org}} where the users rate on a $5$-star system.

The Yahoo Music dataset~\cite{dror2012yahoo} comprises $262,810,175$ ratings of $624,961$ music items by $1,000,990$ users collected during the year $1999-2010$. The ratings include one-minute resolution timestamps, allowing refined temporal analysis. The average sparsity of the whole dataset is about 0.42\%.  This Yahoo Music dataset was collected from the Yahoo Music service.\footnote{\url{https://www.yahoo.com/music/}} The ratings range from $1$ to $100$.



\begin{table} [t]
\caption{Results on MovieLens Dataset}
\label{tbl:performance-ml}
\centering
\begin{threeparttable}
\begin{center}
\begin{tabular}{l||c| c |c | c}
\hline\hline
 & \scriptsize{POMDP-Rec} & \scriptsize{timeSVD++} & \scriptsize{F M} & \scriptsize{M F} \\
\hline
\tiny{$RMSE$}  & {\bf 0.8419}\tnote{$\star$} & 0.8423 & 0.8589 & 0.8519 \\
\hline
\end{tabular}
\begin{tablenotes}
  \scriptsize{\item[$\star$] The variance of the results for POMDP-Rec is $0.0017$}
\end{tablenotes}

\end{center}
\end{threeparttable}
\end{table}

\begin{table} [t]
\caption{Results on Yahoo Music Dataset}
\label{tbl:performance-ym}
\centering
\begin{threeparttable}
\begin{center}
\begin{tabular}{l||c| c |c | c}
\hline\hline
 & \scriptsize{POMDP-Rec} & \scriptsize{timeSVD++} & \scriptsize{F M} & \scriptsize{M F} \\
\hline
\tiny{$RMSE$}  & {\bf 22.769}\tnote{$\star$} & 23.381 & 23.551 & 24.372 \\
\hline
\end{tabular}
\begin{tablenotes}
  \scriptsize{\item[$\star$] The variance of the results for POMDP-Rec is $0.739$}
\end{tablenotes}

\end{center}
\end{threeparttable}
\end{table}

\subsection{Performance comparison}
In order to simulate the real world recommendation scenario, we would like to split the training data and the testing data by time. We use about $80\%$ data from earlier time as the training data, and the remaining data for testing. As discussed in the previous section, the data are split and scattered into time windows, and the predictions of the POMDP-Rec are evaluated for each time windows. We report the average performance of the POMDP-Rec on the MovieLens dataset in Table~\ref{tbl:performance-ml}, and the Yahoo Music dataset in Table~\ref{tbl:performance-ym}.

In order to compare with the previous results, we adopt KDDCUP${}^\prime11$ official evaluation metric, root-mean-square error (RMSE). That is, $RMSE = \sqrt{ \sum_{i=1}^m \sum_{j=1}^n I_{ij} (\bR_{ij} - \hat{\bR}_{ij})^2 / \mid I \mid}$, where $i$, $j$ are indices for users and items respectively, $I_{ij}$ is the $0/1$ indicator function for availability of rating $R_{ij}$.

The baselines include
\begin{itemize}
  \item {\bf timeSVD++}~\cite{koren2009collaborative}. This method is reported to achieve excellent performance on the Netflix dataset~\cite{bennett2007netflix} by considering the time changing behaviors throughout the life span of the data.
  \item {\bf Factorization Machine (FM)}~\cite{rendle2010factorization}. As a general predictor working with any real valued feature vector, FM combines the advantages of Support Vector Machine with factorization models. It achieves good results in several Kaggle (\url{http://www.kaggle.com}) recommendation competitions.
  \item {\bf Matrix Factorization (MF)}. MF is a classical collaborative filtering approach for recommendations. We use the libMF implementation~\cite{zhuang2013fast}.
\end{itemize}
The parameters are all fine-tuned. In order to perform fair comparisons, we fixed the dimension of latent factors in $\bU$ and $\bV$ to be $32$ in all baselines and the POMDP-Rec.

As a reference, for the Yahoo Music dataset, the best ensemble of models achieves $RMSE\approx21.0$, and the best single model achieves $RMSE\approx22.1$~\cite{chen2012linear}, by the official KDDCUP${}^\prime11$ competition report~\cite{dror2012yahoo}. However, the parameters of these winning models are fine-tuned exhaustively by domain experts and the performance depends much on the feature engineering, while our model does not take much human efforts in fitting any specific recommendation mechanisms.

\subsection{Analysis of POMDP-Rec}

\begin{figure}
\centering
\includegraphics[width=0.65\textwidth]{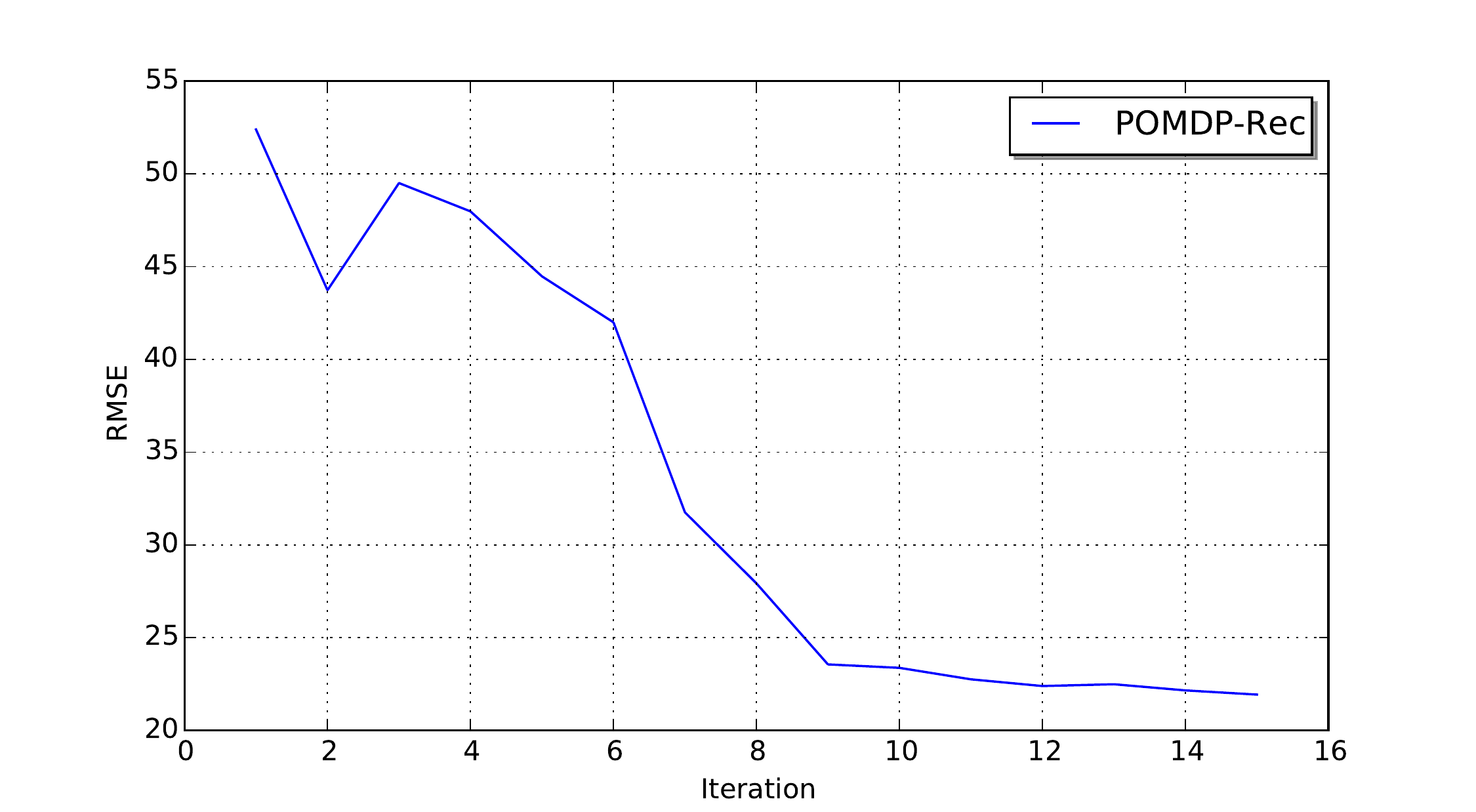}
\caption{The performance of POMDP-Rec improves when the model is trained for more iterations.} \label{fig:update_process}
\end{figure}
In this section, we analyze the POMDP-Rec framework empirically on the Yahoo Music dataset.
Previous literatures~\cite{hasselt2010double} showed that the estimation of the value function could be rather biased, especially when there are large over-estimations of future Q-values. It appears that iterating over the randomly shuffled sample set for multiple times could improve the performance. In the experiments, we evaluate the POMDP-Rec framework after each iteration, and show the trend of performance improvements in Figure~\ref{fig:update_process}.

\begin{figure}[t!]
\centering
\subfigure[Average reward]{\label{fig:stabR}\includegraphics[clip, width=0.45\textwidth]{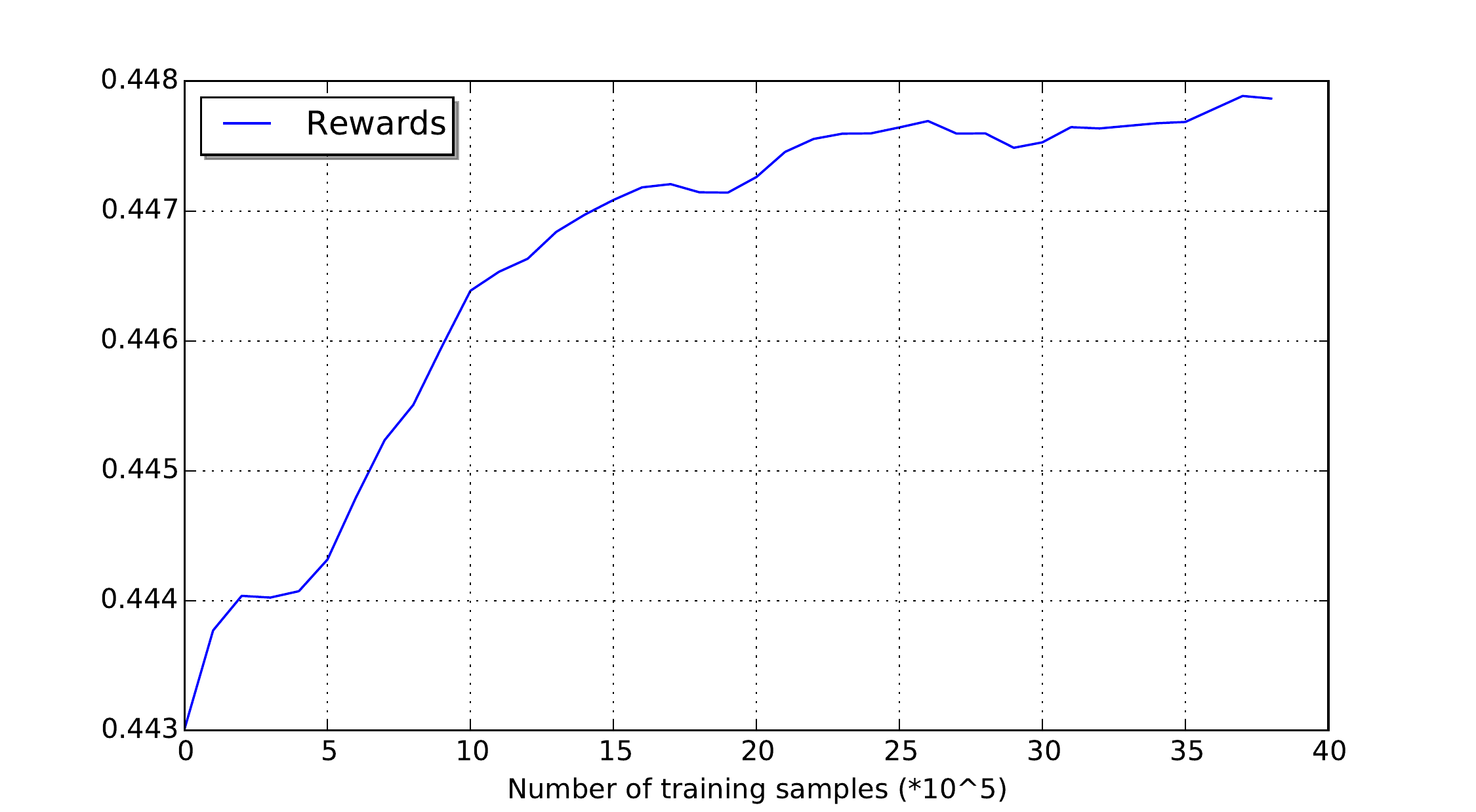}}\hspace{7mm}
\subfigure[Average maximal Q-value]{\label{fig:stabQ}\includegraphics[clip, width=0.45\textwidth]{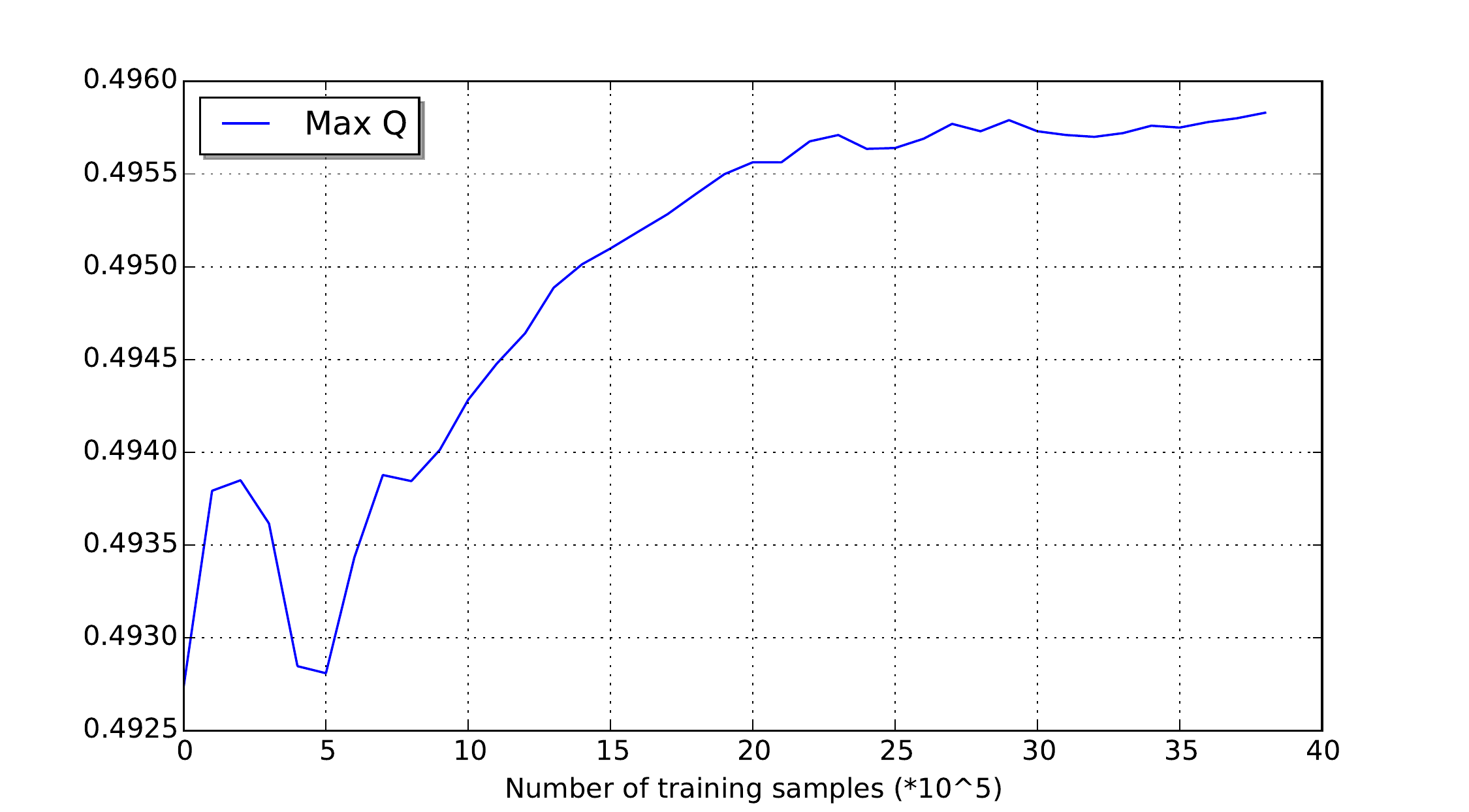}}
\caption{Stability of the POMDP-Rec framework.}\label{fig:stab}
\end{figure}


From Figure~\ref{fig:update_process}, we notice that stating from iteration $6$, the model's performance has a significant improvement. We would like to exam its stability starting from iteration $6$. In the previous researches on Q-learning~\cite{mnih2013playing}, the average reward and the average $\max Q$ are often adopted as indicators of the model stability. Therefore, we plot the average reward and the average $\max Q$ with respect to the number of training samples in Figure~\ref{fig:stab}. The POMDP-Rec framework is stable in modeling over the Yahoo Music recommendation dataset, because we see in Figure~\ref{fig:stabR} the average rewards smoothly increases as the performance of the model improves, and in Figure~\ref{fig:stabQ} the average $\max Q$ appears to be stable (i.e., slope becomes small in the figure) after a large amount of training samples have been provided.


\section{Related Works}\label{section:related}

The sequential nature of the recommendation process was noticed in the past~\cite{shani2005mdp}. The common practice is to model the sequential nature via the Markove Decision Process (MDP)~\cite{shani2005mdp,sahoo2012hidden}. Taking the Markovian idea to model the sequential reccommendations one step further, we suggest the sequential recommendation shall be considered as a Partially Observable MDP. Because in the settings of recommender systems, the user's actions may not be fully observed, but every actions we observed can be used to collaboratively estimate the distributions of the user's interests. As an intuitive sketch of a natural modeling of a recommender system, at each time interval, the recommender system observes users' actions, infers its belief states, and makes recommendations. These processes are seamlessly modeled through our proposed POMDP-Rec framework.

From the perspective of handling the RD phenomenon, our work is also related to solving the one-class problem. As one of the causes of the RD phenomenon in recommender systems, one class problem refers to the recommendation scenario where only the positive feedbacks is received. To solve this problem, there are mainly two approaches. On one hand, the researchers proposed to augment the training data by generating negative samples~\cite{marlin2009collaborative,hernandez2014probabilistic}. On the other hand, the reinforcement learning approaches have been proposed, such as the contextual bandit models~\cite{li2010contextual,yue2012hierarchical}.

In terms of methodology, this work is inspired by the recent advances of deep reinforcement learning researches~\cite{mnih2013playing,mnih2015human,lillicrap2015continuous,hausknecht2015deep}. A deep reinforcement learning model employs a deep network to estimate the value function of each discrete action, and when acting, select the maximally valued output for a given state input. We adopt the parameterized optimization for the value function in our work.

\section{Conclusion}\label{section:conclusion}
We propose the POMDP-Rec framework designed to model the recommendation process. The POMDP-Rec framework handles the changing distribution of the testing samples through transiting between its states. The learning of the POMDP-Rec does not require a balance of the positive and the negative samples. And the Markov property of the POMDP-Rec eliminates the impacts of recurrent training. With these nice properties, the POMDP-Rec framework achieves good results in the off-line experiments on both the MovieLens $1$M dataset and the Yahoo Music dataset. In the future, we would like to deploy the POMDP-Rec framework to a real-world online advertising recommender system, train and evaluate it based on the practical metrics (e.g., CTR or revenue), and try to analyze the learned belief states.

\small
\bibliographystyle{abbrv}
\bibliography{nips_2016}

\end{document}